\documentclass{article}

\usepackage[dblblindworkshop, final]{neurips_2025}
\workshoptitle{The 4th Deep Learning for Code Workshop (DL4C)}

\usepackage[utf8]{inputenc} 
\usepackage[T1]{fontenc}    
\usepackage{hyperref}       
\usepackage{url}            
\usepackage{booktabs}       
\usepackage{amsfonts}       
\usepackage{nicefrac}       
\usepackage{microtype}      
\usepackage{xcolor}         
\usepackage[pdftex]{graphicx}
\usepackage{tabularx}
\usepackage{fontawesome5}
\usepackage{enumitem}

\title{Increasing LLM Coding Capabilities through\\Diverse Synthetic Coding Tasks}

\author{
  Amal Abed$^{1*}$ \quad
  Ivan Lukic$^{1*}$ \quad
  Jörg K.H. Franke$^{1,2,3, 4}$ \quad
  Frank Hutter$^{1,2,5}$ \quad
  \\ \\
  $^1$University of Freiburg \quad
  $^2$ELLIS Institute Tübingen \quad \\
  $^3$Open-Sci Collective \quad
  $^4$LAION \quad
  $^5$Prior Labs \quad
  \\ \\
  \footnotesize{$^*$Equal contribution}
}

\begin{document}

\maketitle



\begin{abstract}
Large language models (LLMs) have shown impressive promise in code generation, yet their progress remains limited by the shortage of large-scale datasets that are both diverse and well-aligned with human reasoning. Most existing resources pair problems with solutions, but omit the intermediate thought process that guides coding.
To close this gap, we present a scalable synthetic data generation pipeline that produces nearly 800k \textit{instruction--reasoning--code--test} quadruplets. Each sample combines a task, a step-by-step reasoning trace, a working solution, and executable tests, enabling models to learn not just the \emph{what} but also the \emph{how} of problem solving. Our pipeline combines four key components: curated contest problems, web-mined content filtered by relevance classifiers, data expansion guided by reasoning patterns, and multi-stage execution-based validation. A genetic mutation algorithm further increases task diversity while maintaining consistency between reasoning traces and code implementations.
Our key finding is that fine-tuning LLMs on this dataset yields consistent improvements on coding benchmarks. Beyond raw accuracy, reasoning-aware data can substitute for model scaling, generalize across architectures, and outperform leading open-source alternatives under identical sample budgets. 
Our work establishes reasoning-centered synthetic data generation as an efficient approach for advancing coding capabilities in LLMs. We publish our dataset and generation pipeline to facilitate further research. 
\footnote{\makebox[1.5em][c]{\raisebox{-0.2\height}{\faGithub}}  
  \href{https://github.com/ivlu2000/diverse-codegen-pipeline}{GitHub Repository} \quad \makebox[2em][c]{\raisebox{-0.2\height}{\includegraphics[height=1.2em]{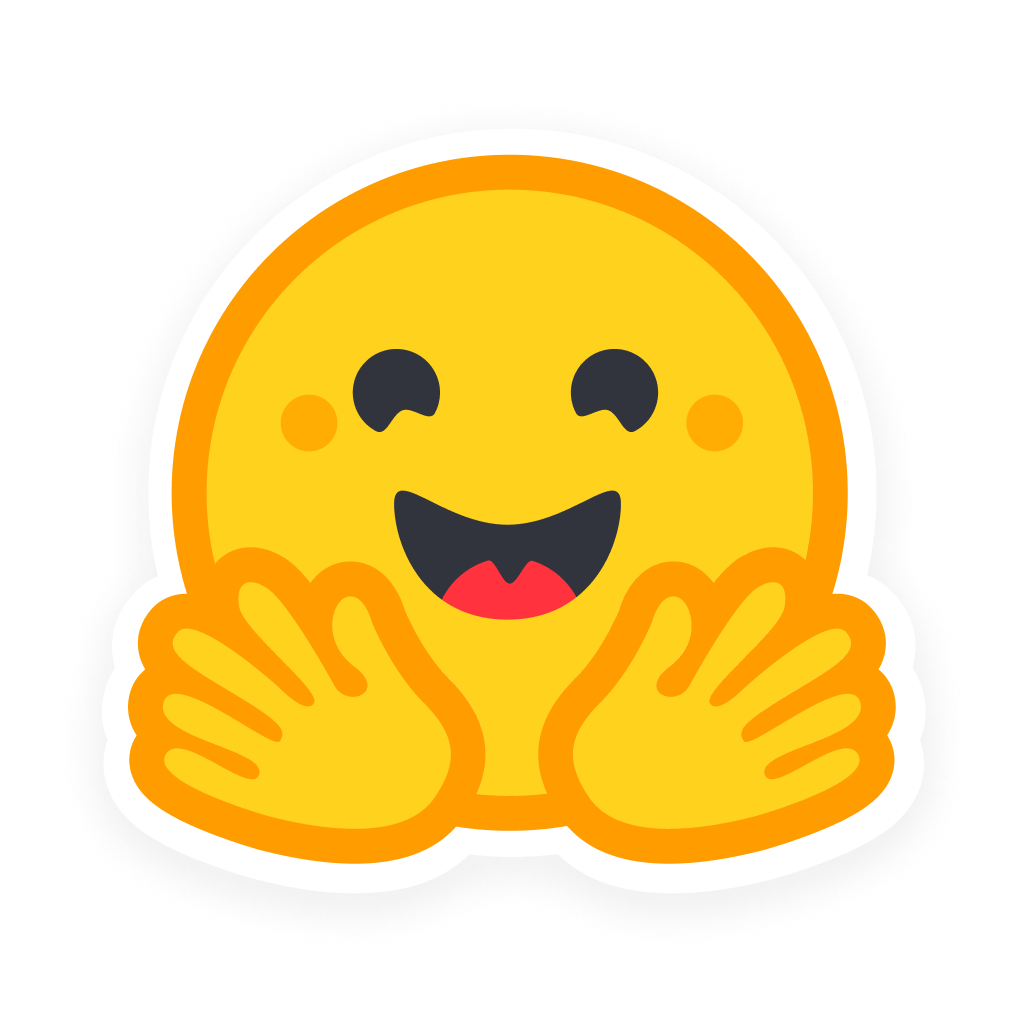}}}  
  \href{https://huggingface.co/datasets/amal-abed/combined_dataset}{Hugging Face Dataset}}
\end{abstract}

\section{Introduction}

Large language models (LLMs) have shown strong progress in code generation \cite{hui2024qwen25codertechnicalreport, 10.1145/3747588}, yet their limitations become clear on tasks requiring systematic reasoning and generalization. While benchmarks such as HumanEval \citep{chen2021evaluating} and MBPP \citep{austin2021mbpp} highlight the potential of scaling, they also expose a persistent bottleneck: the lack of training data that jointly captures \emph{diversity, reasoning, and functional correctness} at scale. Without such resources, models often succeed on familiar problem types but fail to adapt to new challenges or to explain how solutions are derived.

Most available datasets provide instruction–code pairs or final solutions, but omit the intermediate reasoning that connects problem understanding to executable code. This omission matters because reasoning traces offer a training signal that can improve both reliability and interpretability. Human-annotated resources with reasoning exist but are prohibitively expensive to scale, leaving a gap for building models that can both \emph{generate correct programs} and \emph{reveal the logic behind them}. Synthetic datasets have begun to address this, yet many focus narrowly on correctness or rely on costly closed models, limiting both scalability and openness.

To address this gap, we present a reproducible pipeline for constructing synthetic \emph{instruction–reasoning–code–test} datasets designed to advance LLM coding capabilities. Starting from curated seed problems, we expand coverage through corpus filtering, generate structured quadruplets with mid-sized open models, and enforce quality via execution-grounded validation. To further broaden problem coverage, we introduce a genetic instruction mutation algorithm that creates novel task variants without sacrificing correctness.

The resulting dataset captures not only solutions but also the logical traces behind them, offering a scalable resource for training and evaluation. Controlled comparisons against leading open-source datasets show that our formulation consistently delivers stronger transfer performance under identical budgets, underscoring the importance of reasoning-augmented and diversity-driven data generation.

\section{Related Work}

Recent efforts have focused on scaling dataset construction for code generation. \textit{Self-Code-Align} \citep{wei2024selfcodealign} filters raw functions into structured coding concepts for controlled task creation, while \textit{Magicoder} \citep{wei2024magicoder} and \textit{InverseCoder} \citep{wu2024inversecoderunleashingpowerinstructiontuned} leverage open-source code to produce instruction–code pairs or invert the mapping to generate instructions from code. Other approaches expand task complexity through evolution, as in \textit{WizardCoder} \citep{luo2024wizardcoder} and \textit{EpiCoder}  \citep{wang2025epicoder}, or through biologically inspired mutation and crossover, as in \textit{Genetic Instruct} \citep{majumdar-etal-2025-genetic}.  

While these pipelines improve scale and diversity, they rarely capture intermediate reasoning or systematically enforce correctness. Our work extends this line by generating \textbf{instruction–reasoning–code–test quadruplets}, validated through execution and guided by classifier-based filtering, yielding datasets that couple semantic diversity with functional reliability.

\section{Methodology}
We propose an end-to-end methodology for constructing large-scale, reasoning-augmented datasets for code generation. The pipeline is designed to transform heterogeneous programming material into standardized, validated, and diverse problem sets that directly support model training and evaluation. The process begins with broad task collection and expansion, balancing curated contest-style problems with large-scale mining of real-world programming sources. These raw tasks are then systematically structured into instruction–reasoning–solution–test quadruplets, ensuring that each sample exposes both problem statements and associated reasoning traces. A rigorous execution-based validation stage enforces functional correctness, discarding faulty generations while preserving diversity through multi-candidate refinement. To scale beyond limited seed material, we integrate a genetic-inspiration framework that iteratively evolves tasks through crossover and mutation, yielding novel but coherent programming challenges. Finally, a multi-stage deduplication process safeguards against redundancy, while fine-tuning and benchmark evaluations quantify efficiency, generalization, and dataset impact. By unifying curation, structuring, validation, expansion, and deduplication into a single cohesive workflow, our methodology balances scale, reliability, and reasoning fidelity in dataset construction.

\subsection{Dataset Curation and Expansion}
Our pipeline begins with a seed collection of curated programming tasks in the \citep{leetcode} style—\textit{title, description, constraints, and examples}—which are particularly useful because they support automatic test generation. However, the publicly available set of only $\sim$2.3k such problems is insufficient for training large language models. To broaden coverage, we incorporated tasks from competitive programming platforms such as \citep{codeforces} and \citep{atcoder}, combining existing HuggingFace datasets with custom scraping pipelines. These sources introduce greater diversity in problem structure and difficulty, though they often lack reference implementations. Consequently, reasoning traces, candidate code, and executable tests were systematically generated in later stages of our pipeline.

To scale further and move beyond the limited design space of curated contest tasks, we employed a classifier-guided mining approach inspired by ~\citet{shao2024deepseekmath}. Specifically, we trained a FastText \citep{bojanowski2017enriching} classifier on the curated problems and applied it to the 3B-document DCLM-Baseline corpus \citep{li2024datacomp}, a high-quality subset of  Common Crawl\citep{commoncrawl}. By enforcing a strict 90\% relevance threshold, we extracted $\sim$4M candidate documents, striking a balance between high recall of coding-related material and effective filtering of irrelevant or noisy content. This step allowed us to capture a much broader distribution of real-world programming challenges, ranging from algorithmic puzzles to applied coding snippets, thereby providing a richer substrate for the subsequent reasoning-augmented generation stages.The resulting mix of curated and mined material forms the substrate for structured transformation with LLMs.

\subsection{Structuring into Instruction–Reasoning–Solution–Test Quadruplets}
From the pool of curated and mined documents, we employed \textbf{Qwen2.5-Coder-7B-Instruct} \citep{hui2024qwen25codertechnicalreport} (Apache 2.0 MIT License) to convert raw programming content into standardized \textit{instruction–reasoning–solution–test} quadruplets. This stage is critical: raw problems from contest archives or web mining are often unstructured, ambiguous, or lack consistent interfaces. The LLM first reformulates each problem into a clear, self-contained instruction, then generates step-by-step reasoning traces that connect problem statements to code implementations, followed by \textbf{three candidate solution–test pairs}. This standardized format ensures that downstream models are exposed not only to final answers but also to the intermediate reasoning strategies that support generalization.

We selected Qwen2.5-Coder-7B-Instruct for its balance of efficiency and capability. Its moderate size makes it practical for large-scale generation while still providing strong reasoning and coding abilities. In addition, it is widely available and well-documented, which supports reproducibility and ease of integration into our pipeline. While larger variants such as Qwen2.5-Coder-32B \citep{hui2024qwen25codertechnicalreport} can offer higher quality, their computational cost is prohibitive at scale. Importantly, our pipeline remains model-agnostic: different LLMs can be substituted to trade efficiency for further quality improvements without altering the overall process.

\subsection{Execution-Based Validation and Refinement}
Building on prior execution-validated dataset pipelines such as MAmmoTH2~\cite{yue2024mammoth2} and Self-Code-Align~\cite{wei2024selfcodealign}, we incorporated a rigorous multi-candidate refinement stage to ensure that generated samples were both reliable and functionally correct. For each candidate instruction, the LLM was tasked with producing a reasoning trace alongside three alternative \textit{solution–test} pairs. These solutions were executed inside isolated Python containers with strict limits on runtime, memory, and external calls, thereby preventing unsafe or non-terminating code. This multi-candidate approach substantially reduced the risk of discarding otherwise valid problems due to a single poor generation.

The validation process selected the first solution that passed all corresponding test cases, ensuring consistency between reasoning, implementation, and execution. Samples for which no candidate passed were discarded, preventing propagation of faulty code. Beyond correctness, this stage acted as a powerful filter against hallucinated reasoning traces or malformed test cases, refining the dataset into a coherent, executable form.

With this validation in place, the resulting dataset provided a reliable foundation for subsequent scaling and augmentation, which we address next with the Genetic-Instruct framework.

\subsection{Evolutionary Expansion with Genetic-Instruct}

\begin{figure}[h]
    \centering
    \includegraphics[width=0.8\textwidth]{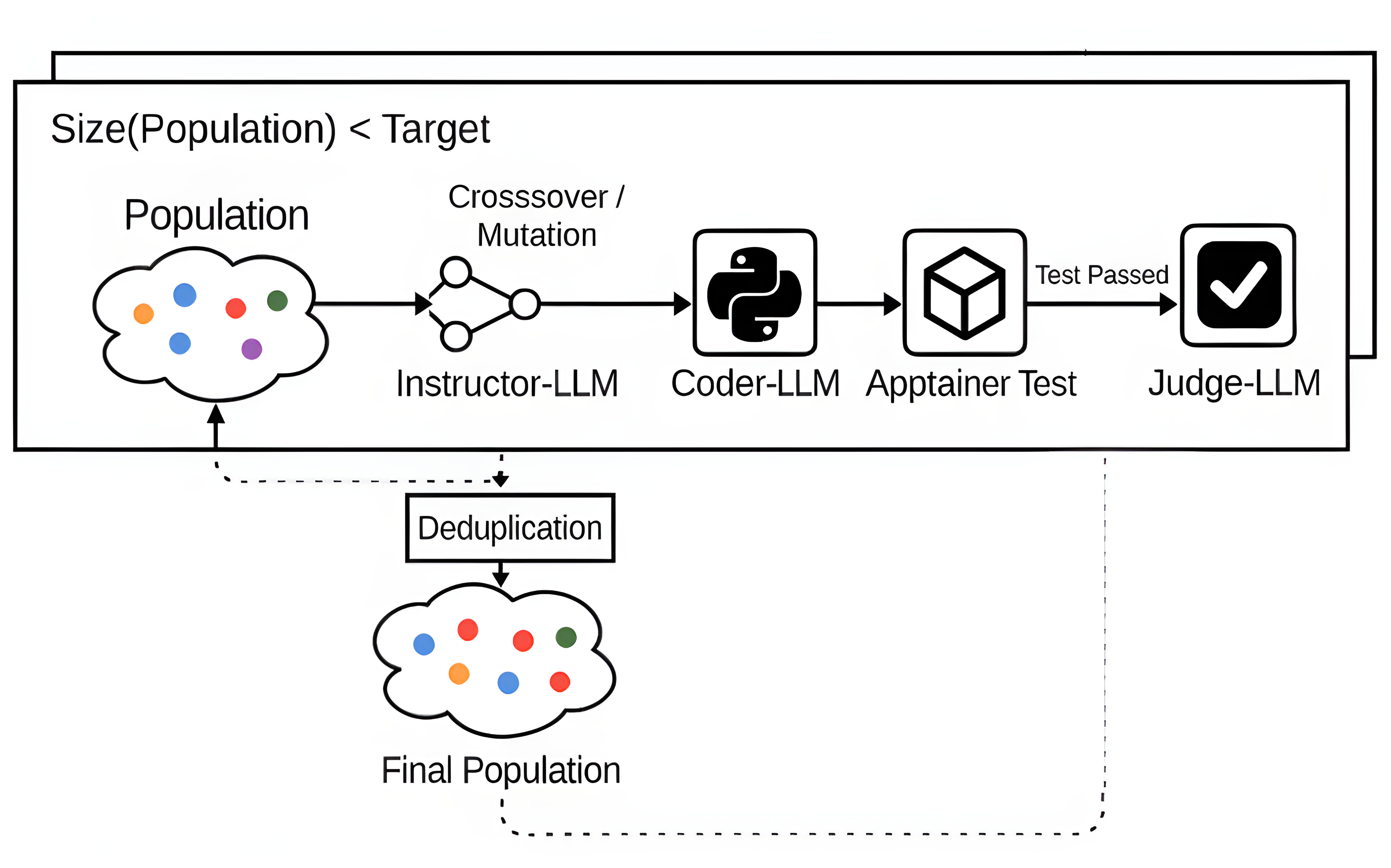}
    \caption{Process Flow for Genetic Instruct}
    \label{fig:llm_process}
\end{figure}

To scale beyond the initial seed set, we adopted a \textit{Genetic-Instruct} framework \citep{majumdar-etal-2025-genetic}, which iteratively evolves new tasks from existing ones (Figure~\ref{fig:llm_process}). This design is inspired by genetic algorithms: instead of generating problems entirely from scratch, the system reuses validated instructions as a population and applies controlled transformations to increase diversity while preserving internal consistency. Each cycle operates on a pool of high-quality instructions and produces a new generation of candidate tasks that inherit structure and reasoning patterns from their predecessors.

At the core of each cycle, the \textit{Instructor-LLM} generates not only a new instruction but also its accompanying reasoning trace. Two complementary operators guide this process:  
\begin{itemize}[leftmargin=1em, itemsep=0pt]
    \item \textbf{Crossover:} The LLM is given five seed tasks as few-shot examples and instructed to synthesize exactly one new instruction by merging elements from at least two of them. The resulting hybrid task inherits aspects such as constraints, objectives, or reasoning strategies from multiple parents. Crucially, the Instructor-LLM also produces a coherent reasoning trace that integrates these elements, ensuring the derived problem is both novel and logically solvable.  
    \item \textbf{Mutation:} The LLM perturbs an individual task through prompt-driven transformations. Mutations include tightening or adding constraints, increasing reasoning depth, or expanding the problem scope. As with crossover, each mutated task is paired with a fresh reasoning trace that remains consistent with the modified instruction.  
\end{itemize}

The generated \textit{instruction–reasoning} pairs then proceed through the rest of the pipeline. The \textit{Coder-LLM} produces three candidate \textit{solution–test} pairs for each task, guided by the reasoning trace. These are parsed into scripts and executed in secure Apptainer containers under strict resource limits, ensuring isolation, safety, and reproducibility. The first candidate that passes all tests is selected as the final implementation; if none succeed, the instruction is discarded.  

Finally, the \textit{Judge-LLM} enforces structural, semantic, and functional quality. In addition to verifying formatting and clarity, it checks that the generated code indeed solves the instruction in alignment with the reasoning trace. Only tasks that pass this judgment are retained. To sustain both quality and diversity, the instruction pool is periodically refreshed: every 200k accepted tasks, the validated set is mixed with the seed pool before entering the next cycle. This iterative feedback loop prevents collapse into repetitive formulations and steadily expands the dataset into a broader space of programming challenges while maintaining correctness and reasoning fidelity.

\subsection{Deduplication, Diversity Preservation, and Decontamination}
A critical challenge in large-scale dataset construction is avoiding redundancy, since repeated or near-identical problems can bias training and inflate benchmark performance. To mitigate this, we applied a multi-stage deduplication pipeline. First, all instructions were embedded using \textbf{MiniLM-L6-v2}\citep{all-MiniLM-L6-v2}, a lightweight but effective sentence-transformer model. We then performed approximate nearest-neighbor search with \textbf{FAISS} \citep{douze2024faiss} to efficiently identify candidate duplicates across the hundreds of thousands of structured problems in our dataset. Pairs with cosine similarity above 0.90 were flagged for further inspection.
Because surface-level similarity does not always imply true duplication (e.g., two sorting problems with different constraints), flagged pairs were passed to a locally hosted \textbf{Gemma-3-27B-IT} \citep{team2025gemma} model for verification. This LLM-based verifier judged whether two instructions were functionally identical, even if phrased differently or containing minor variations. Confirmed duplicates were merged using a \textit{union--find} clustering procedure, which groups all linked pairs into equivalence classes and retains only one representative per class. This approach ensured that the final dataset preserved diversity in problem formulation while eliminating redundancy both in surface phrasing and in underlying functionality.

In addition to deduplication, we also conducted a thorough data leakage check against common code evaluation benchmarks, namely \textbf{HumanEval} and \textbf{MBPP}. To this end, we computed and compared hashes of benchmark problems with those in our dataset. The comparison revealed \textbf{zero overlap}, confirming that our dataset does not contain leaked benchmark problems and is suitable for reliable downstream fine-tuning and evaluation.

\subsection{Fine-Tuning Setup}  
To rigorously measure the impact of our dataset, we fine-tuned \textbf{Phi-2}, a 2.7B-parameter transformer developed by Microsoft \citep{phi2_model} (MIT License). Phi-2 was selected because it strikes a favorable balance between scale and capability: despite its relatively modest size compared to recent large language models, it demonstrates strong reasoning and code generation ability. This makes it a cost-effective and informative testbed, allowing us to isolate the contributions of our dataset without confounding factors introduced by extremely large model architectures.  

Fine-tuning was carried out using the QLoRA \citep{NEURIPS2023_1feb8787} framework, which enables efficient adaptation of large models by applying low-rank updates to a subset of parameters while keeping the majority of the model frozen. Specifically, we set the rank to $r=16$, the scaling factor to $\alpha=16$, and targeted the modules \texttt{[q\_proj, v\_proj, k\_proj, dense]}. Training was conducted for 10 epochs, providing sufficient exposure to the dataset for the model to adapt to the reasoning-augmented patterns while remaining computationally feasible.

\subsubsection{Evaluation}  
We assessed the fine-tuned models on two widely used benchmarks: \textit{HumanEval}, which emphasizes algorithmic reasoning and problem-solving, and \textit{MBPP}, which contains shorter programming tasks with more direct mappings from instructions to solutions. To ensure consistency and rigor in evaluation, we adopted the \texttt{EvalPlus} framework \citep{liu2023is}, which extends the original benchmark test suites with additional cases and more robust correctness checks. This offers a stronger measure of generalization beyond the limited canonical test sets.

\section{Results}  
\label{sec:results}

\subsection{HumanEval and MBPP Benchmarks}

The base \texttt{phi-2 2.7B} achieved 45.7\% (Base) and 40.9\% (Extra) on HumanEval, and 62.7\% / 51.6\% on MBPP. LeetCode \citep{greengerong2023leetcode} fine-tuning offered only marginal improvements, underscoring the limitations of small, curated datasets. In contrast, our synthetic data consistently boosted performance, with gains that scaled with dataset size. At 25k synthetic samples, pass rates reached 56.1\% / 51.8\% on HumanEval and 65.6\% / 55.3\% on MBPP—representing nearly +10 absolute points over baseline on HumanEval.  

\begin{table}[t]
  \caption{Pass rates (\%) on HumanEval and MBPP for \texttt{phi-2 2.7B}}
  \label{tab:phi2_results}
  \centering
  \begin{tabularx}{\linewidth}{l *{4}{>{\centering\arraybackslash}X}}
    \toprule
    \textbf{Model} & \multicolumn{2}{c}{\textbf{HumanEval}} & \multicolumn{2}{c}{\textbf{MBPP}} \\
    \cmidrule(lr){2-3} \cmidrule(lr){4-5}
     & Base Test (\%) & Extra Tests (\%) & Base Test (\%) & Extra Tests (\%) \\
    \midrule
    Base Model (Phi-2 2.7B)        & 45.7 & 40.9 & 62.7 & 51.6 \\
    Fine-tuned on LeetCode dataset & 47.6 & 42.1 & 63.0 & 51.6 \\
    Fine-tuned on 5k synthetic samples  & 54.3 & 49.4 & 64.3 & 54.5 \\
    Fine-tuned on 10k synthetic samples & 54.9 & 50.6 & 65.6 & 55.3 \\
    Fine-tuned on 25k synthetic samples & \textbf{56.1} & \textbf{51.8} & \textbf{65.6} & \textbf{55.3} \\
    \bottomrule
  \end{tabularx}
\end{table}

Overall, these experiments confirm that synthetic, reasoning-augmented data offers a scalable and effective path toward improving coding performance across diverse benchmarks.  

\subsection{Efficiency Gains}

\begin{figure}[t]
    \centering
    \includegraphics[width=1.0\textwidth]{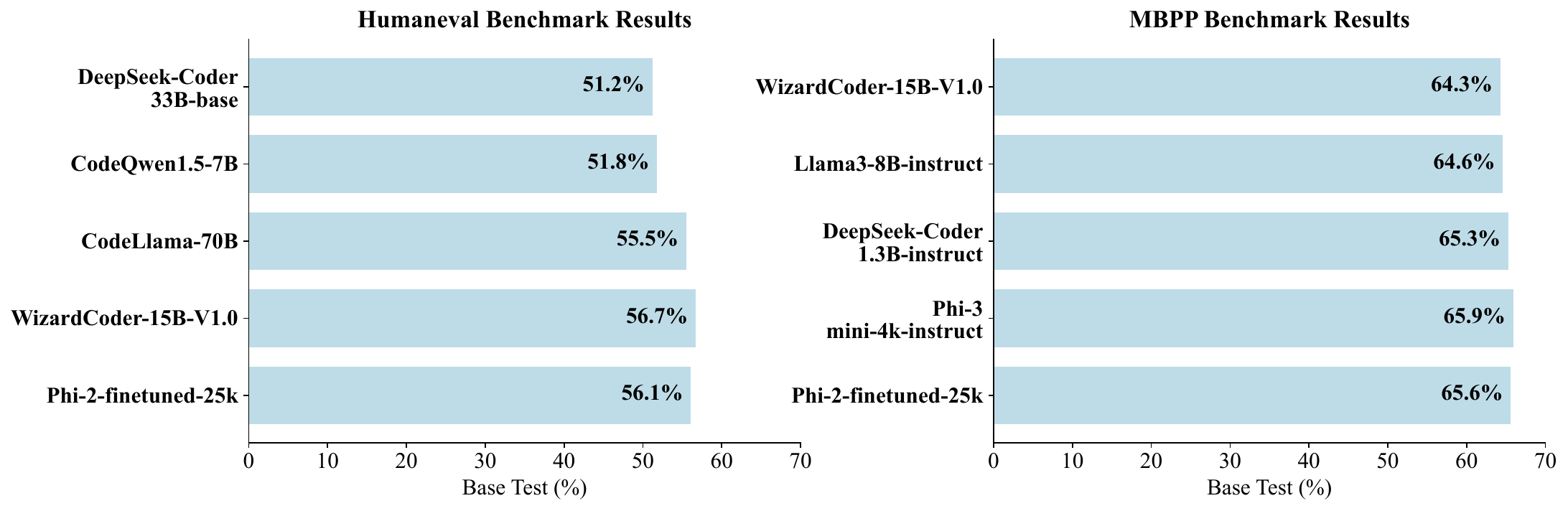}
    \caption{Pass rates on HumanEval and MBPP benchmarks for selected models.}
    \label{fig:models}
\end{figure}

Synthetic, reasoning-augmented fine-tuning proved to be a more efficient alternative to scaling model size. The fine-tuned \texttt{phi-2 2.7B} achieved competitive, and in some cases superior, performance compared to substantially larger models such as \texttt{CodeLlama-70B} \citep{roziere2023code}, \texttt{Llama3-8B-instruct} \citep{dubey2024llama}, and \texttt{DeepSeek-Coder-33B-base} \citep{guo2024deepseek} (Figure~\ref{fig:models}). This demonstrates that targeted synthetic data can significantly narrow the performance gap between small and large models, offering a more compute-efficient path to progress.  

The implications extend beyond raw benchmark numbers. Scaling to tens of billions of parameters requires specialized hardware, large-scale distributed training, and high inference costs, all of which limit accessibility. In contrast, our approach shows that carefully curated, reasoning-focused data allows models an order of magnitude smaller to deliver comparable gains. This not only reduces training and deployment costs, but also broadens access to competitive code generation systems for researchers and practitioners with modest resources.  

In this context, efficiency means achieving strong performance without the prohibitive expenses of brute-force scaling. Synthetic, reasoning-augmented fine-tuning therefore emerges as a practical and sustainable alternative, advancing the capabilities of code generation models while keeping training and inference requirements manageable.

\section{Experiments}

Our experiments test whether targeted, reasoning-augmented fine-tuning can improve code generation without simply scaling model size. We evaluate along two complementary axes: (1) \textbf{generalization}, by assessing whether improvements transfer across architectures such as \texttt{CodeGemma-2B}; and (2) \textbf{dataset quality}, by contrasting homogeneous versus diverse subsets of our data and benchmarking against other recent open-source resources. Together, these studies isolate the contributions of scale, architecture, and dataset design.

To ensure fair comparisons, we standardized fine-tuning using QLoRA. A one-epoch configuration search was conducted over eight variants, varying rank (\( r \in \{16,32\} \)), scaling (\( \alpha = 2r \)), target modules (\texttt{[q\_proj, v\_proj]} or \texttt{[q\_proj, v\_proj, k\_proj]}), and whether to retain the final classification head (\texttt{save\_head} $\in \{\texttt{True}, \texttt{False}\}$). The best-performing configuration on downstream benchmarks was then applied consistently across all models and datasets.

\textbf{Compute resources.} All fine-tuning experiments were run on a single NVIDIA A100 GPU (80GB) using QLoRA. Each configuration search run completed within $\sim$1–2 hours, while fine-tuning for 10 epochs with the selected hyperparameters required up to $\sim$12 hours for the largest dataset (25k examples). We report results from single-GPU runs to ensure reproducibility and accessibility, avoiding reliance on large-scale distributed infrastructure.

\subsection{Cross-Model Generalization}

To evaluate whether these gains are architecture-specific, we applied the same training subsets to \texttt{CodeGemma-2B} \citep{team2024codegemma}(Gemma License), a model already specialized for coding. For fair comparison, we re-fine-tuned \texttt{phi-2} under identical conditions (same subsets, training setup, and evaluation protocol) to those used for \texttt{CodeGemma}. As shown in Table~\ref{tab:codegemma_results}, CodeGemma benefited strongly from our dataset: fine-tuning on 25k samples yielded a +14.6 point increase on HumanEval base tests (23.2\% $\rightarrow$ 37.8\%) and a +6.8 point gain on MBPP base tests (55.6\% $\rightarrow$ 62.4\%), with comparable improvements on extra tests. Interestingly, the scaling behavior was consistent with that of \texttt{phi-2}, suggesting that our dataset delivers distinct benefits even for models already optimized for code.

\begin{table}[t]
  \caption{Pass rates (\%) on HumanEval and MBPP for \texttt{CodeGemma-2B}}
  \label{tab:codegemma_results}
  \centering
  \begin{tabularx}{\linewidth}{l *{4}{>{\centering\arraybackslash}X}}
    \toprule
    \textbf{Model} & \multicolumn{2}{c}{\textbf{HumanEval}} & \multicolumn{2}{c}{\textbf{MBPP}} \\
    \cmidrule(lr){2-3} \cmidrule(lr){4-5}
     & Base Test (\%) & Extra Tests (\%) & Base Test (\%) & Extra Tests (\%) \\
    \midrule
    Base Model (CodeGemma-2B)       & 23.2 & 17.7 & 55.6 & 45.8 \\
    Fine-tuned on LeetCode          & 31.1 & 25.0 & 57.7 & 48.4 \\
    Fine-tuned on 5k synthetic      & 33.5 & 27.4 & 60.6 & 50.0 \\
    Fine-tuned on 10k synthetic     & 36.6 & 31.1 & \textbf{62.4} & \textbf{52.6} \\
    Fine-tuned on 25k synthetic     & \textbf{37.8} & \textbf{32.3} & \textbf{62.4} & 51.6 \\
    \bottomrule
  \end{tabularx}
\end{table}

Taken together, these findings highlight two important aspects: (1) reasoning-augmented synthetic data enables efficient specialization of smaller models, reducing reliance on costly scaling, and (2) the benefits generalize across architectures and pretraining regimes. Both points strengthen the case for dataset-driven approaches as a practical alternative to simply increasing model size. Having established that our data scales efficiently across models, we next ask whether \emph{the structure of the dataset itself}—in particular, its diversity versus homogeneity—affects downstream performance.

\subsection{Diversity vs. Homogeneity}

\begin{figure}[t]
    \centering
    \includegraphics[width=0.7\textwidth]{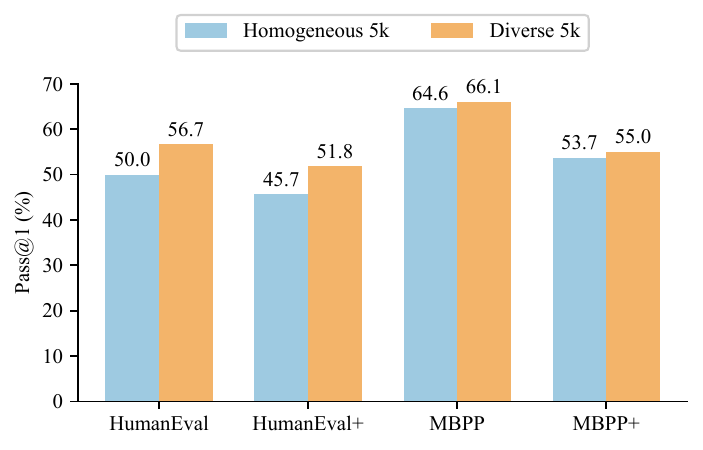}
    \caption{Pass@1 scores on HumanEval and MBPP for homogeneous vs. diverse subsets. Diversity yields stronger reasoning alignment.}
    \label{fig:homogeneous_experiment}
\end{figure}

To disentangle the effects of scale and coverage, we compared matched 5k subsets under two conditions: \emph{homogeneous} and \emph{diverse}. The homogeneous subsets were sampled from our dataset \emph{before the deduplication step}, resulting in highly similar, single-domain style problems. In contrast, the diverse subsets were drawn from the final deduplicated dataset, covering a broad range of domains and reasoning types.  

Despite being identical in size, the diverse subsets consistently outperformed the homogeneous ones. On HumanEval, diverse fine-tuning reached 56.7\% compared to 50.0\% for homogeneous, and 51.8\% versus 45.7\% on HumanEval+. On MBPP, the effect was smaller but still present: 66.1\% vs.\ 64.6\%, and 55.0\% vs.\ 53.7\% on MBPP+. These results highlight that redundancy and narrow domain focus reduce the effective information content of training data, while exposure to a variety of problem formulations improves downstream generalization.  

Taken together, these findings show that at small and medium training budgets, \emph{diversity is more valuable than raw sample count}. Even relatively small but heterogeneous datasets can rival or exceed larger, less varied ones, underscoring the importance of deduplication and domain coverage in building effective resources for code generation. While these results highlight benefits within code-focused benchmarks, it remains essential to confirm that domain-specific fine-tuning preserves broader reasoning ability. We turn to this in the next section.

\subsection{General Reasoning Benchmarks}
A common concern with domain-specific fine-tuning is that it might reduce a model’s broader reasoning ability. To verify this, we evaluated our models on three standard benchmarks outside programming: HellaSwag (commonsense inference) \citep{Zellers2019HellaSwagCA}, WinoGrande (coreference reasoning) \citep{10.1145/3474381}, and MMLU (multi-task knowledge)\citep{Hendrycks2020MeasuringMM}. Across all three, accuracy remained essentially unchanged, with variations well within normal fluctuation (see Table~\ref{tab:eval_results_phi2_style}).

\begin{table}[t]
  \caption{Evaluation results (\%) across HellaSwag, WinoGrande, and MMLU for base, 5k-fine-tuned, and 25k-fine-tuned models}
  \label{tab:eval_results_phi2_style}
  \centering
  \begin{tabularx}{\linewidth}{l *{3}{>{\centering\arraybackslash}X}}
    \toprule
    \textbf{Dataset / Metric} & \textbf{Base Model} & \textbf{5K Fine-tuned} & \textbf{25K Fine-tuned} \\
    \midrule
    HellaSwag (Acc)       & 55.88 & 55.22 & 55.50 \\
    HellaSwag (Acc\_Norm) & 73.76 & 73.00 & 73.32 \\
    WinoGrande (Acc)      & 75.93 & 76.09 & 76.01 \\
    MMLU (Avg Acc)        & 54.37 & 53.94 & 54.14 \\
    \bottomrule
  \end{tabularx}
\end{table}

These results indicate that reasoning-aware code fine-tuning preserves general capabilities. In other words, models can gain substantial improvements in code generation without sacrificing the versatility needed for non-coding tasks, making this approach suitable for building specialized yet broadly capable systems. Finally, to contextualize our approach against existing resources, we compare our dataset directly with other recent open-source alternatives.

\subsection{Comparison with Other Datasets}

\begin{figure}[t]
    \centering
    \includegraphics[width=1.0\textwidth]{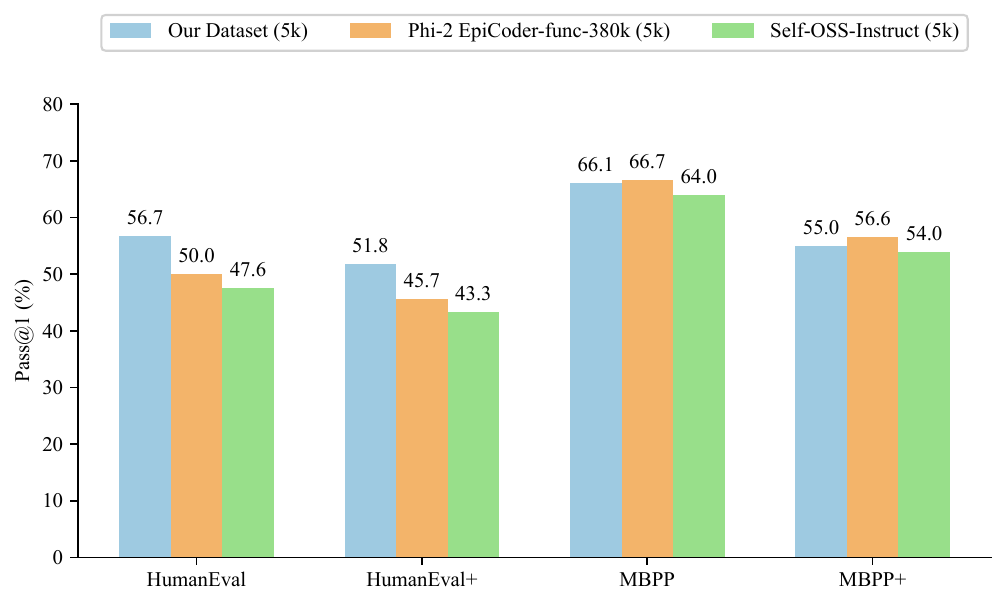}
    \caption{Comparison of pass@1 scores on HumanEval and MBPP for models fine-tuned on 5k samples from our dataset versus EpiCoder-func-380k and Self-OSS-Instruct-SC2-Exec-Filter-50k.}
    \label{fig:dataset_experiment}
\end{figure}

To put our dataset in context, we benchmarked it against two recent open-source resources: \textit{EpiCoder-func-380k} \citep{wang2025epicoder}(MIT License) and \textit{Self-OSS-Instruct-SC2-Exec-Filter-50k} \citep{wei2024selfcodealign}. For fairness, we fine-tuned on matched 5k subsets from each source under identical training settings. Across both HumanEval and MBPP, models trained on our data consistently achieved higher pass rates, even though all experiments used the same sample budget (Figure~\ref{fig:dataset_experiment}).  

The comparisons also highlight where our approach makes the biggest difference. Improvements were most pronounced on HumanEval, which contains more complex programming tasks that often require multi-step reasoning. This suggests that our dataset’s emphasis on reasoning-oriented problem formulations, genetic instruction variation, and rigorous validation is particularly valuable for benchmarks that go beyond surface-level coding skills. While MBPP also showed gains, they were smaller, consistent with its focus on simpler problems. Together, these results indicate that careful dataset design can yield benefits beyond scale alone, especially when the downstream tasks demand structured reasoning.

\section{Conclusion}

This work demonstrates that large-scale, reasoning-augmented synthetic datasets can play a decisive role in advancing the coding capabilities of large language models. By combining FastText-based corpus filtering, Qwen2.5-Coder generation, multi-stage validation, and our \textit{Genetic Instruct} algorithm for systematic instruction variation, we produced nearly \textbf{800k high-quality instruction–reasoning–code–test quadruplets}. Unlike existing resources, our dataset captures not just solutions, but also the intermediate reasoning processes that enable models to generalize to harder problems. The resulting data proved to be diverse, interpretable, and reliable—qualities that our experiments identified as critical drivers of performance.

Extensive evaluations highlight the strength of this approach. Fine-tuning on our synthetic subsets consistently surpassed strong baselines such as LeetCode, and even allowed smaller models to rival or outperform substantially larger ones. These results underline that the key to progress is not raw dataset size, but the integration of reasoning and diversity. Further, cross-model experiments confirmed that the benefits generalize across architectures, while evaluations on general reasoning tasks showed no degradation—demonstrating the robustness of our pipeline.

A current limitation of our work is that the pipeline supports only Python. Extending it to other programming languages will require adapting the execution-based validation framework, which is essential to ensure functional correctness in more diverse coding environments.

Taken together, our findings establish synthetic, diversity-driven data generation as a powerful and scalable foundation for building more capable and efficient code-focused LLMs. Looking forward, extending this pipeline to multilingual programming languages (e.g., Java, C++, JavaScript) and deploying it at pretraining scale offers a clear pathway toward models that can reason, generalize, and adapt across programming paradigms. In a landscape where real-world datasets are limited, our results highlight that it is the \emph{breadth and structure of data}—not raw scale—that unlocks the next generation of robust, reasoning-capable models for code.

\newpage

\section*{Acknowledgements}

This research was funded by the Deutsche Forschungsgemeinschaft (DFG, German Research Foundation) under grant number 417962828.
We acknowledge funding by the European Union (via ERC Consolidator Grant DeepLearning 2.0, grant no.~101045765). Views and opinions expressed are, however, those of the author(s) only and do not necessarily reflect those of the European Union or the European Research Council. Neither the European Union nor the granting authority can be held responsible for them. \begin{center}\includegraphics[width=0.3\textwidth]{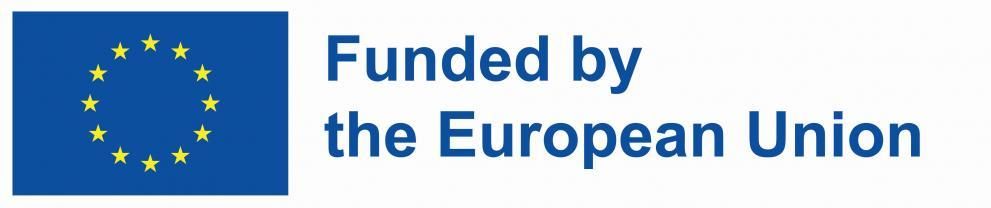}\end{center}

We gratefully acknowledge the Gauss Centre for Supercomputing e.V. for funding this work by providing computing time through the John von Neumann Institute for Computing (NIC) on the supercomputer JUWELS Booster at Jülich Supercomputing Centre (JSC).

\bibliographystyle{plainnat}
\bibliography{references}

\end{document}